\relax
\documentclass[letterpaper]{article} 
\usepackage{aaai21}  
\usepackage{times}  
\usepackage{helvet} 
\usepackage{courier}  
\usepackage[hyphens]{url}  
\usepackage{graphicx} 
\usepackage{natbib}  
\usepackage{caption} 
\usepackage{hyperref}
\usepackage{dirtree}
\title{KD-Lib: A PyTorch library for Knowledge Distillation, Pruning and Quantization}
\author{

    Het Shah,\textsuperscript{\rm1}
    Avishree Khare,\textsuperscript{\rm2}\thanks{Equal contribution}
    Neelay Shah,\textsuperscript{\rm3}$^*$
    Khizir Siddiqui \textsuperscript{\rm4}$^*$
}

\affiliations{

    \{f20170093\textsuperscript{\rm 1}, f20170112\textsuperscript{\rm 2}, f20180400\textsuperscript{\rm 3}, f20180439\textsuperscript{\rm 4}\}@goa.bits-pilani.ac.in

}

\begin{document}

\maketitle


\begin{abstract}
    In recent years, the growing size of neural networks has led to a vast amount of research concerning compression techniques to mitigate the drawbacks of such large sizes. Most of these research works can be categorized into three broad families : Knowledge Distillation, Pruning, and Quantization. While there has been steady research in this domain, adoption and commercial usage of the proposed techniques has not quite progressed at the rate. We present KD-Lib, an open-source PyTorch based library, which contains state-of-the-art modular implementations of algorithms from the three families on top of multiple abstraction layers. KD-Lib is model and algorithm-agnostic, with extended support for hyperparameter tuning using \href{https://optuna.org/}{Optuna} and \href{https://www.tensorflow.org/tensorboard}{Tensorboard} for logging and monitoring.
\linebreak
    The library can be found at - \url{https://github.com/SforAiDl/KD_Lib}
\end{abstract}

\section{Introduction}
Deep neural networks (DNNs) have gained widespread popularity in recent years, finding use in several domains including computer vision, natural language processing, human computer interaction and more. 
These networks have achieved remarkable results on several tasks, often even surpassing human-level performance.

The number of parameters of such DNNs often increase multi-fold with an increase in their representation capacity, limiting the deployment capabilities and hence, the commercial feasibility of these networks. This limitation warrants the need for efficient compression techniques that can shrink the networks in size while ensuring that the drop in performance is minimal. In this paper, we restrict our focus to three widely-used compression techniques: Knowledge Distillation, Network Pruning and Quantization.

Knowledge Distillation \cite{Hinton2015} is a compression paradigm that leverages the capability of large neural networks (called teacher networks) to transfer knowledge to smaller networks (called student networks).  While large models (such as very deep neural networks or ensembles of many models) have higher knowledge capacity than small models, this capacity might not be fully utilized. It can be computationally just as expensive to evaluate a model even if it utilizes little of its knowledge capacity. Knowledge distillation aims to transfers knowledge from a large model to a smaller model without loss of validity. Several advancements have been witnessed in the development of richer knowledge distillation algorithms, attempting to reduce the difference in test accuracies of the teacher and the student. These algorithms are model-agnostic and hence can be used for a wide variety of network architectures.

While knowledge distillation attempts to train an equally-competent smaller network, network pruning \cite{lecun1990} attempts to reduce the size of the existing network by removing unimportant weights. Different pruning techniques differ in the choice of weights to eliminate and the methods used to do the same. Pruning can help in reducing the size of the network up to 90\% with minimal loss in performance. Some approaches have also been empirically shown to result in faster training of the pruned network along with a higher test accuracy \cite{frankle2018lottery}. 

Quantization is another way to compress neural networks by reducing the number of bits used to store the weights. As the weights of a network are usually stored as 32-bit floating values (FP32), reducing the precision to 8-bit integer values (INT8) will reduce the size of the network by 4 times. Several approaches have been developed to quantize networks with minimal loss in performance.   

These compression techniques have become extremely popular in recent years and are actively being researched. New algorithms proposed in research papers can be difficult to understand and implement, especially for potential users in a non-academic setting, thereby limiting their commercial usage. To the best of our knowledge, there does not exist an umbrella framework containing implementations of state-of-the-art algorithms in Knowledge Distillation, Pruning and Quantization. In this paper, we present KD-Lib, a comprehensive PyTorch based library for model compression. KD-Lib aims to bridge the gap between research and widespread use of model compression techniques. We envision that such a framework would be helpful to researchers as well, providing them a tool to build upon existing algorithms and helping them in going from idea to implementation faster.

\begin{table*}[!htb]
\centering
\begin{tabular}{|l|l|l|l|}
\hline
\textbf{Library}                     & \textbf{Knowledge Distillation}                 & \textbf{Pruning} & \textbf{Quantization}  \\
\hline
KD-Lib (Ours)                        &           Present                               &   Present        &      Present   \\
\hline
Distiller\cite{nzmora2019distiller} &           Present (only 1 algorithm)             &   Present        &       Present  \\
\hline
AIMET\textsuperscript{\rm3}         &           -                                     &   Present        &      Present   \\
\hline
AquVitae\textsuperscript{\rm1}      &           Present                               &   -              &        -       \\
\hline
Distiller\textsuperscript{\rm1}     &           Present                               &   -              &        -       \\
\hline
\end{tabular}
\caption{Comparision of various libraries with KD-Lib}
\label{tab:rel-works}
\end{table*}

\section{Related work}

We compare KD-Lib with several openly available frameworks and libraries. In our comparison, we do not include libraries that support less than two algorithms. 
\bigskip

Distiller \cite{nzmora2019distiller} is the most extensive framework we found, but it primarily focuses on quantization and pruning with only one knowledge distillation algorithm \cite{Hinton2015}. AquVitae\footnote{\url{https://github.com/aquvitae/aquvitae}} contains 4 distillation methods but no quantization and pruning algorithms. Similarly Distiller\footnote{\url{https://github.com/karanchahal/distiller}} has 11 knowledge distillation techniques but lacks pruning and quantization methods. AIMET\footnote{\url{https://github.com/quic/aimet}} focuses mainly on quantization and some other relatively less popular model compression techniques such as tensor decomposition. \linebreak
In our survey, we found no library containing algorithms pertaining to all 3 of the popular compression paradigms - knowledge distillation, pruning and quantization. 
Table \ref{tab:rel-works} shows concise comparison with different frameworks.

\section{Features and Algorithms} 

KD-Lib houses several algorithms proposed in recent years for model compression. The following features have driven the design choices for the library: 
\begin{itemize}
    \item The main aim of KD-Lib is to make model compression algorithms accessible to a wide range of users, and hence the work is fully open-source.
    \item The library should act as a catalyst for further research in these fields. It should also be extendable to newer algorithms and other model compression fields. Hence, it is designed to be modular, allowing flexible modifications to essential components that can lead to novel algorithms or better extensions to existing algorithms. 
    \item The interface should be easy to use. Hence, the core functionalities (distillation/pruning/quantization) are accessible in a few lines of code. 
    \item As tuning the hyperparameters is essential for optimum performance, KD-Lib provides support for hyperparameter tuning via Optuna. Monitoring and logging support is also provided through Tensorboard.
\end{itemize} 
\noindent
A brief description of the implemented algorithms is as follows: 
\begin{itemize}
    \item \textbf{Knowledge Distillation :} The algorithms have been divided into two major task-types: Vision and Text. The Vision module currently supports 13 algorithms while the Text module supports distillation from BERT to LSTM-based networks\cite{BERT2LSTM2019}.
    \item \textbf{Pruning :} The library currently supports pruning based on the Lottery ticket Hypothesis \cite{frankle2018lottery}.
    \item \textbf{Quantization :} Static Quantization, Dynamic Quantization and Quantization Aware Training (QAT) \cite{QAT2018} are currently supported by KD-Lib. 
\end{itemize}

\section{Code Structure}
The structure of the library has been designed for efficient use with the following major principles kept in mind:

\begin{itemize}
    \item The core function of an algorithm can be executed in one line of code. Hence, the classes contain a dedicated method for distillation/pruning/quantization.
    \item Each module allows extension to newer features and easy modifications. Hence, fluid components of algorithms (loss functions in distillation, for example) can be easily customized. 
    \item Necessary statistics are available wherever needed. Hence, methods dedicated to these are also present ($get\_pruning\_statics$, for example). 
\end{itemize}

\begin{figure}[!htb]
    \centering
    \begin{minipage}{\textwidth}
        \dirtree{%
            .1 Distiller.
            .2 \textit{train\_student}.
            .2 \textit{train\_teacher}.
            .2 \textit{evaluate}.
            .2 \textit{calculate\_kd\_loss}.
        }
    \end{minipage}
    \caption{Structure of a Distiller.}
    \label{fig:fig_distill}
\end{figure}

Knowledge Distillation algorithms can be accessed as Distiller objects (Figure 1), with at least the mentioned methods. The \textit{train\_student} method distills knowledge from a teacher network to a student network, where the teacher network could optionally be trained using the \textit{train\_teacher} method. The \textit{evaluate} method can be invoked to test the performance of the student network. The \textit{calculate\_kd\_loss} method can overridden to provide a custom loss function for distillation. This can also be leveraged by researchers to test novel Knowledge Distillation loss functions.  

\begin{figure}[!htb]
    \centering
    \begin{minipage}{\textwidth}
        \dirtree{%
            .1 Pruner.
            .2 \textit{prune}.
            .2 \textit{get\_pruning\_statistics}.
        }
    \end{minipage}
    \caption{Structure of a Pruner.}
    \label{fig:fig_prune}
\end{figure}

Pruning algorithms have been implemented as Pruner objects (Figure 2). Each Pruner object can access the \textit{prune} method for pruning the network. Additionally, the \textit{get\_pruning\_statistics} method can be used to obtain information about the weights of the network after pruning (percentage of network pruned, for example). 

\begin{figure}[!htb]
    \centering
    \begin{minipage}{\textwidth}
        \dirtree{%
            .1 Quantizer.
            .2 \textit{quantize}.
            .2 \textit{get\_performance\_statistics}.
            .2 \textit{get\_model\_sizes}.
        }
    \end{minipage}
    \caption{Structure of a Quantizer.}
    \label{fig:fig_quant}
\end{figure}

Quantization algorithms can be accessed via Quantizer objects (Figure 3). The \textit{quantize} method can be used for quantization (with differing implementations for different algorithms). Additionally, the \textit{get\_model\_sizes} method can be used to compare sizes of the model before and after quantization and the \textit{get\_performance\_statistics} method can be used to compare test-times and error metrics for the two networks.

\noindent
The documentation for the library\footnote{\url{https://kd-lib.readthedocs.io/}} has the description of all classes and selected tutorials with example code snippets.

\section{Benchmarks}

We summarize benchmark results on some of the algorithms implemented in KD-Lib in Tables \ref{tab:kd-table}, \ref{tab:pruning-table} and \ref{tab:qaunt-table}.

\begin{table}[!htbp]
\centering
\begin{tabular}{|c|c|c|c|}
\hline
\textbf{Algorithm} & \textbf{Accuracy} \\
\hline
None  & 0.57 \\
\hline
DML \cite{DML}  & 0.62              \\
\hline
Self Training \cite{selfknowledge}  & 0.61              \\
\hline
Messy Collab \cite{nosiycollab}  & 0.60             \\
\hline
Noisy Teacher \cite{noisyteacher}  & 0.59             \\
\hline
TAKD \cite{TAKD}  & 0.59              \\
\hline
RCO \cite{routeconstrained}  & 0.58              \\
\hline
Probability Shift \cite{KA}  & 0.58              \\
\hline
\end{tabular}
\caption{ The accuracies of networks trained by some of various knowledge distillation algorithms KD-Lib packages on the CIFAR10 dataset. All models were trained with the same hyperparameter set to ensure a fair comparison. We consider ResNet34 as the teacher network (with an accuracy of 0.63) and report accuracies for the student network (ResNet18). \emph{None} refers to a ResNet18 model trained from scratch without any model compression algorithm. The compression ratio for all of the knowledge distillation algorithms is 50.7\% }
\label{tab:kd-table}
\end{table}

\begin{table}[!htbp]
\centering
\begin{tabular}{|l|l|l|}
\hline
\textbf{Pruning Epoch} & \textbf{\% Model Pruned} & \textbf{Accuracy} \\
\hline
1                      & 0.0                      & 0.9878  \\
\hline
2                      & 0.10                     & 0.9891 \\
\hline
3                      & 0.19                    & 0.9890 \\
\hline
\end{tabular}
\caption{Pruning percentage and accuracy of ResNet18 model on MNIST using Lottery Ticket Pruning \cite{frankle2018lottery}. Each pruning epoch consists of 5 training epochs. 'Model pruned' is the percentage of model pruned and 'Accuracy' is the corresponding accuracy at the end of the epoch.}
\label{tab:pruning-table}
\end{table}

\begin{table}[!htbp]
\centering
\begin{tabular}{|c|c|c|c|}
\hline
\textbf{Algorithm} &  \textbf{\% Size Change} & \textbf{BA} & \textbf{NA}\\
\hline
Static  & -0.75 & 0.72        & 0.70              \\
\hline
QAT & -0.75 & 0.72 & 0.71 \\
\hline
Dynamic & -0.19 & 0.70        & 0.70            \\
\hline
\end{tabular}
\caption{Comparison of various quantization algorithms. 'BA' (Base Accuracy) is the accuracy of the model before quantization, and 'NA' (New Accuracy) is the accuracy of the model after quantization. '\% Size change' refers to the change in size after quantization. In Static Quantization and QAT, ResNet18 is tested on the CIFAR10 dataset. For Dynamic Quantization, LSTM is tested on IMDB dataset.}
\label{tab:qaunt-table}
\end{table}


\section{Conclusion and Future Work}

In this paper, we present KD-Lib, an easy-to-use PyTorch-based library for Knowledge Distillation, Pruning and Quantization. KD-Lib is designed to facilitate the adoption of current model compression techniques and act as a catalyst for further research in this direction. We plan on actively maintaining the library and also expanding it to include more algorithms and desirable features (distributed training, for example) in the future. We further plan on extending this library to other domains relevant to the research community including but not limited to explainability and interpretability in knowledge distillation. 

\begin{quote}
\begin{small}
\bibliography{ref.bib}
\end{small}
\end{quote}

\end{document}